\documentclass[onecolumn]{webofc}

\usepackage[varg]{txfonts}   
\usepackage{hyperref}       
\usepackage{subcaption}
\usepackage{url}            
\begin{document}
\title{Reactor Optimization Benchmark by Reinforced Learning}
%
%

\author{\firstname{Deborah} \lastname{Schwarcz}\inst{1}\fnsep\thanks{\email{deborahsc@soreq.gov.il}} \and
         \firstname{Nadav} \lastname{Schneider}\inst{1,3}\fnsep\thanks{\email{nadavsch@post.bgu.ac.il}} \and
        \firstname{Gal} \lastname{Oren}\inst{2,4}\fnsep\thanks{\email{galoren@cs.technion.ac.il}} \and
        \firstname{Uri} \lastname{Steinitz}\inst{1}\fnsep\thanks{\email{urist@soreq.gov.il}}
}

\institute{Soreq Nuclear Research Center, Israel 
\and Scientific Computing Center, Nuclear Research Center -- Negev, Israel
\and Department of Electrical \& Computer Engineering, Ben-Gurion University, Israel
\and Department of Computer Science, Technion -- Israel Institute of Technology}

\abstract{%
    Neutronic calculations for reactors are a daunting task when using Monte Carlo (MC) methods. As high-performance computing has advanced, the simulation of a reactor is nowadays more readily done, but design and optimization with multiple parameters is still a computational challenge. MC transport simulations, coupled with machine learning techniques, offer promising avenues for enhancing the efficiency and effectiveness of nuclear reactor optimization. This paper introduces a novel benchmark problem within the OpenNeoMC framework designed specifically for reinforcement learning. The benchmark involves optimizing a unit cell of a research reactor with two varying parameters (fuel density and water spacing) to maximize neutron flux while maintaining reactor criticality. The test case features distinct local optima, representing different physical regimes, thus posing a challenge for learning algorithms. Through extensive simulations utilizing evolutionary and neuroevolutionary algorithms, we demonstrate the effectiveness of reinforcement learning in navigating complex optimization landscapes with strict constraints. Furthermore, we propose acceleration techniques within the OpenNeoMC framework, including model updating and cross-section usage by RAM utilization, to expedite simulation times. Our findings emphasize the importance of machine learning integration in reactor optimization and contribute to advancing methodologies for addressing intricate optimization challenges in nuclear engineering. The sources of this work are available at our GitHub repository: {\href{https://github.com/Scientific-Computing-Lab-NRCN/RLOpenNeoMC}{RLOpenNeoMC}}.}
\maketitle

\section{Monte Carlo Optimizations for Nuclear Reactor Design}

The synergy between Monte Carlo (MC) transport simulations and machine learning techniques has emerged as a promising avenue for advancing nuclear reactor calculations. With the increasing scale and complexity of these simulations, there is a pressing need for efficient optimization methods to enhance computational performance. One notable endeavor in this direction is the development of the OpenNeoMC framework \cite{gu2023openneomc}, which integrates evolutionary, reinforcement, or neuroevolutionary optimization techniques with MC simulations. By harnessing the capabilities of the neutronic transport code OpenMC \cite{romano2013openmc} and the algorithmic framework NeoRL \cite{radaideh2021neorl}, OpenNeoMC offers a streamlined approach to solving intricate optimization problems in reactor physics. Recent studies have demonstrated the effectiveness of this approach, particularly in tasks such as optimizing detector placements within nuclear reactor cores using reinforcement learning techniques \cite{gu2023openneomc}. This integration not only accelerates MC simulations but also holds promise for significantly improving the overall efficiency of nuclear reactor design and analysis.

As with many optimization problems, applying learning algorithms to MC simulations in reactor physics involves sampling a parameter space, where each parameter set corresponds to a calculated value known as the fitness value which is derived from the objective function. The algorithm aims to find parameters that optimize this objective function, typically by seeking the maximum or minimum. However, several distinctive factors in MC simulations of reactor physics necessitate careful consideration:

\begin{itemize}
    \item \textit{Stiff constraints}: The stable operation of a nuclear reactor is done always at critical conditions (a multiplication factor of unity). Even if at realistic conditions the reactivity is regulated by internal and external feedback mechanisms, an optimization of the objective function should reflect this stiffness, and the formulation of the function may substantially affect the convergence of the algorithm.
    \item \textit{Stochasticity}: Being a statistical tool, MC simulation results always feature some variance or 'statistical noise', and as a result, repeated calculations of the same physical model provide different answers. This deviation may be reduced by longer calculations, but can never be eliminated. Thus, as an optimization algorithm zeros in on a solution, it will always wander around within the statistical deviation, and without changing the computation parameters or combining the results of several simulations, it will not be able to converge in the usual sense. 
    \item \textit{Discrete and continuous parameters}: Nuclear reactor optimization may include both discrete parameters (such as core layout and fuel rod selection and refueling scheme) and continuous parameters (fuel density or geometry). There is an advantage to a framework that handles the optimization of both parameter types. 
    \item \textit{Uncertainties}: Any reactor design should incorporate safety features that are robust enough to maintain safe operation within an envelope of uncertainties and tolerances of the actual conditions and the realized construction features. Optimizing the design to accommodate such robustness requires special means and objective functions.
    \item \textit{Burnup and change}: As the reactor operates at full power, the fuel undergoes burnup, and the material changes- the fuel depletes and the density of neutron-absorbing fission products increases. Extension of the optimization to the full reactor lifetime requires calculating the material evolution and adapting the objective function to reflect the continuous operation maximization required.
    \end{itemize}
    
Given the complex challenges inherent in nuclear reactor optimization, the OpenNeoMC framework emerges as a promising solution, particularly in the realm of reinforced learning. Its versatility in accommodating various learning algorithms and its capability to formalize intricate objective functions renders it well-suited for addressing the diverse issues mentioned earlier. Notably, OpenNeoMC is adept at handling both discrete and continuous parameters, with benchmark problems available for each type of parameter search. Moreover, its integration with OpenMC not only accelerates calculations but also facilitates the treatment of burnup problems through specialized modules.

In this study, we present a detailed adaptation of the OpenNeoMC framework and discuss our efforts in accelerating OpenMC calculations using various strategies. These strategies include parallel runs, selective updating of input parameters for repeated calculations, and efficient utilization of RAM for cross-section data storage. We then discuss and compare several optimization methodologies that OpenNeoMC supports, specifically evolutionary and neuroevolutionary algorithms. Leveraging the acceleration methods, we successfully reproduced previous benchmark problems at significantly increased speeds  \cite{gu2023openneomc}.

Furthermore, we introduce a novel physical problem from the realm of nuclear reactor physics, with an objective function that has two distinct local minima. The problem involves an infinite lattice of fuel plates (unit cell), with light water serving as a moderator and cadmium absorbers interspersed between adjacent plate columns. Employing the OpenNeoMC framework, we sought to identify the uranium density and plate spacing configurations that maximize fast neutron flux while adhering to the criticality constraint (multiplication factor $k=1$). In contrast to previous known benchmarks for reactor optimization with OpenNeoMC, this benchmark's two non-connected local optima create an inherent obstacle for reinforcement learning. This makes it suited for testing and benchmarking different learning algorithms and parameters with a physical reactor problem, along with the above-mentioned considerations. 

Through our experimentation with OpenNeoMC, we observed that different learning algorithms for this test case produce qualitatively different results. This observation stresses the importance of selecting appropriate learning algorithms and fine-tuning objective functions to effectively address reactor optimization challenges. Thus, this test case serves as a valuable tool for evaluating and comparing learning algorithms, as well as for exploring strategies to tailor objective functions to specific optimization problems.

\begin{figure}[!h]
    \centering
    \includegraphics[width=0.9\linewidth]{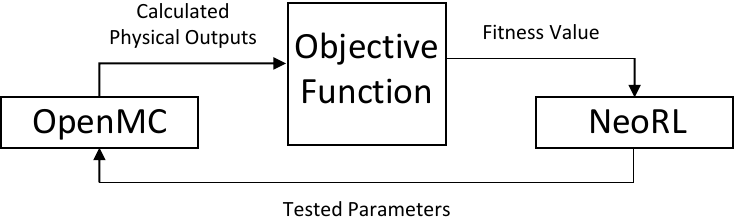}
      \caption{A description of OpenNeoMC's flow, presented in clockwise order: First, OpenMC is invoked with initial parameters within the permissible range of the model. The OpenMC results are received by the objective function which computes the fitness value. The fitness value is then passed through the NeoRL framework, which updates the physical parameters through the chosen algorithm that OpenMC recalculates as the process is repeated.}
      \label{geometrical_description}
\end{figure}

\section{Optimization Methodology}

Using OpenNeoMC (\autoref{openneomc}), two optimization algorithms have been tested. An evolution (JAYA, \autoref{jaya}) and a neuroevolution (PPO-ES, \autoref{ppoes}) algorithms. Each has its advantages and limitations. 

\subsection{OpenNeoMC Framework and its Speed-up}
\label{openneomc}

The OpenNeoMC framework serves as a link between the neutronic transport code OpenMC \cite{romano2013openmc} and the algorithmic framework NeoRL through the implementation of an objective function. A typical OpenNeoMC code comprises three main components (as presented in \autoref{geometrical_description}):

\begin{enumerate}
    \item \textit{Physical Model Definition in OpenMC}: The first component involves defining the physical model in OpenMC, encompassing the system's geometry and materials. This model incorporates one or more physical parameters that may vary within a permissible range of values.
    \item \textit{Objective Function}: The second component entails defining an objective function tailored to the optimization problem at hand. This function uses the calculated OpenMC output to quantify the fitness or the proximity of a solution to the desired goal.
    \item \textit{NeoRL Framework}: The last component involves the integration of the chosen optimization algorithm implemented by NeoRL. This algorithm utilizes the distance to the goal obtained from the objective function and generates the next set of parameters for fitness evaluation. The framework integrates several optimization algorithms (such as evolutionary, reinforcement, or neuroevolutionary), facilitating a comparative study of different algorithms in different settings.
\end{enumerate}

To expedite simulations, parallel optimization is commonly employed in OpenNeoMC \cite{radaideh2021neorl}. Additionally, we have developed a function that updates an existing OpenMC model instead of constructing a new model for each OpenMC simulation, making subsequent runs with nearly similar inputs more efficient. This approach takes into account that only portions of the model change with each simulation, therefore avoiding rebuilding the entire model every time. Furthermore, we allocate RAM for the cross-sections library to speed up the process of retrieving information from libraries. These changes improved the running time by almost $60\%$, of which transferring the cross-section files to RAM contributed the most significant improvement, as demonstrated in \autoref{tab:cost}. These speed-ups were made possible by the open-source nature of OpenMC, as are other performance optimizations \cite{fridman2024distributed}.

\begin{table} [!h]
    \centering
    \caption{Relative running time in different speed improvements of the OpenNeoMC, compared to the initial code. We checked both the improvement in the time duration due to the updating of the model function and in addition to the allocation of RAM to cross-section libraries.}
    \begin{tabular}{|c|c|c|c|} \hline 
         &  Updating model & Updating model + XS in RAM\\ \hline 
         relative running time &  87\% & 43\% \\ \hline
    \end{tabular}
    \label{tab:cost}
\end{table}

   

\subsection{Evolutionary Algorithms and JAYA}
\label{jaya}

Genetic algorithms (GA), inspired by Darwin's theory of natural selection, serve as the foundation for many optimization algorithms. They operate by evolving populations through natural operators such as crossover, mutation, and selection, giving rise to various evolutionary (EA) and swarm algorithms. Notable examples include the particle swarm algorithm (PSO), differential evolution (DE), JAYA, and ant colony optimization (ACO) \cite{shopova2006basic, kennedy1995particle, storn1997differential, rao2016jaya, dorigo1999ant}. The flow of genetic algorithms is presented in \autoref{fig:GAsRL}.

JAYA stands out as an effective EA for solving both constrained and unconstrained optimization problems. Its guiding principle is to move towards the best solution while avoiding the worst solution in the population. JAYA's efficacy is influenced by hyperparameters such as initial population size, number of design variables, and termination criterion \cite{rao2016jaya}.

There have been some attempts to apply genetic algorithms in reactor optimization. For example, Kumar \textit{et al}.\  utilized GA to optimize parameters in the design of a gas-cooled fast breeder reactor core, including thermal-hydraulics analysis and energy transfer \cite{kumar2015new}. 

\subsection{Neuroevolution Algorithms and PPO-ES}
\label{ppoes}

Neural networks, while established long before evolutionary algorithms, gained practical applicability with increased computational power and data availability \cite{schmidhuber2015deep}. However, assembling labeled training datasets for real-life problems remains challenging and costly due to its high dimensionality and dynamic nature.

\begin{figure}[!h]
    \centering
    \includegraphics[width=0.7\linewidth]{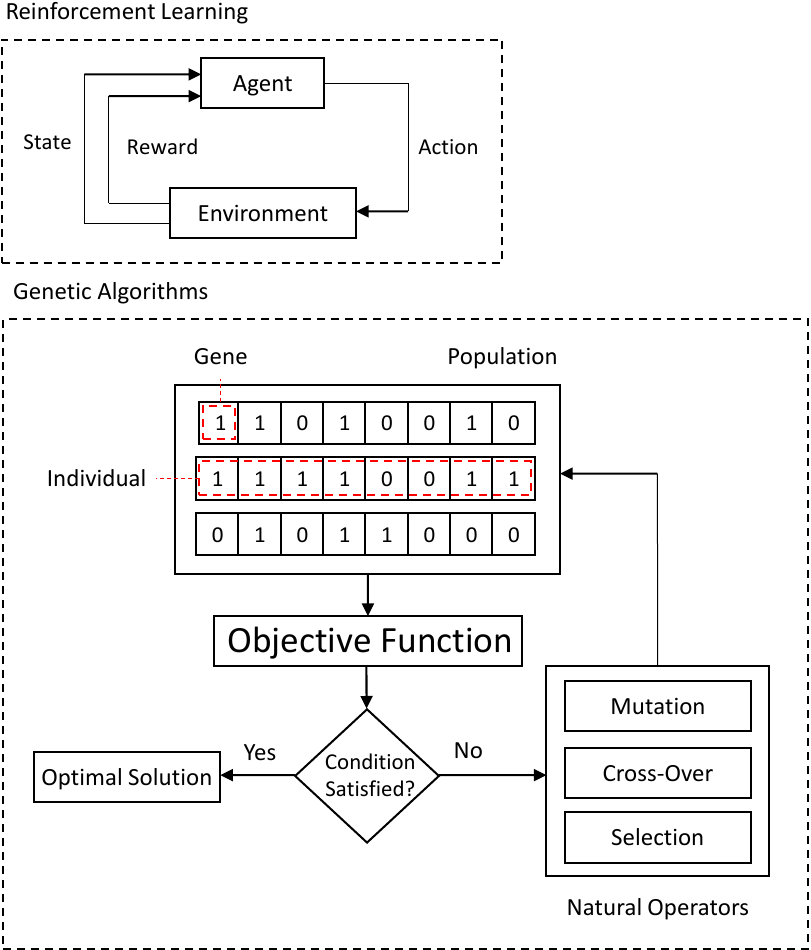}
      \caption{RL trains an agent to react through an environment with rewards given by a chosen objective function. GA and its derivatives are based on an initial population that transforms over generations through natural operators and fitness value given by an objective function until satisfied fitness is achieved. The optimal solution will be the fittest individual.}
      \label{fig:GAsRL}
\end{figure}

Reinforcement Learning (RL) has emerged as a solution, enabling agents to learn sequential decision-making by interacting with environments and maximizing cumulative rewards over time \cite{mnih2015human}. This process is done repetitively until convergence, which implies the agent learned the problem (\autoref{fig:GAsRL}). RL finds applications in nuclear reactor analysis, including prediction of plant behavior and optimization of fuel assemblies for improved efficiency and safety \cite{el2021artificial, radaideh2021physics}.

Neuroevolution algorithms optimize neural networks using evolutionary algorithms, adapting hyperparameters, architectures, and learning algorithms \cite{stanley2019designing}. PPO-ES, a neuroevolution algorithm, combines Proximal Policy Optimization (PPO) with Evolution Strategies (ES) to guide genetic algorithms or evolution strategies (GAs/ES) in large-scale nuclear assembly optimization \cite{radaideh2021large}.

PPO-ES utilizes PPO's policy gradient method with ES's advanced mutation strategies, balancing policy updates with a clip function that constrains the objective function \cite{schulman2017proximal, beyer2002evolution}. Unlike traditional genetic algorithms, ES adapts the mutation strength of individual attributes during the search, enhancing optimization effectiveness. PPO-ES employs PPO to guide ES by running PPO updates in the inner loop of an ES optimization step, perturbing PPO models randomly \cite{liang2017ray}.

\subsection{Benchmarks}
To check our implementation we used two benchmarks of reactor optimization. 
Firstly we use the BEAVRS benchmark model \cite{pin_cell_git} of multiple (unit cell with reflective boundaries) fuel pins with $UO_2$ coated by Zirconium and immersed in borated water. Our goal was to find the enrichment of $UO_2$ to get a multiplication factor of $k=1.1$. We tested this model with two algorithms, JAYA and PPO-ES (using NeoRL framework), and found in both cases similar results to those obtained in the literature \cite{rao2016jaya}. This model has one parameter and a single local (and global) optimum.

In the second benchmark, the goal was to find an optimal configuration of 61 fuel pins in a square construction of an 11×11 grid that maximizes the multiplication factor $k_{eff}$ \cite{k_assembly, k_eff_git}. In this case, we were able to achieve a similar value of $k_{eff}$ as in \cite{k_eff_git}. This benchmark demonstrates the ability to perform optimizations in a discrete parameter set.

These benchmarks have distinct characteristics, the first benchmark has a continuous range of options since the $UO_2$ enrichment can be increased continuously. While in the second benchmark, the configuration of the fuel pin can be modified discretely.

\section{The MTR Reactor RL-Optimization Benchmark}
\label{problem}
While evolutionary algorithms have demonstrated their effectiveness in reactor design optimization \cite{gu2023openneomc}, the exploration of reinforcement learning algorithms in this domain remains underrepresented in existing benchmarks. Unlike evolutionary algorithms, which excel in scenarios with a single global minimum, reinforcement learning algorithms offer distinct advantages in environments characterized by multiple local minima or maxima.

In this paper, we introduce a benchmark system specifically designed to compare reinforcement learning algorithms against their evolutionary counterparts. By highlighting scenarios where the complexity of the optimization landscape, we aim to provide valuable insights into the comparative strengths of these algorithmic approaches in reactor optimization. The system's simplicity both in the physical description and in the constraints, while being related to realistic reactor design challenges, makes it a good benchmark candidate.

\subsection{Physical Description}
\begin{figure}[!ht]
    \centering
    \includegraphics[width=1\linewidth]{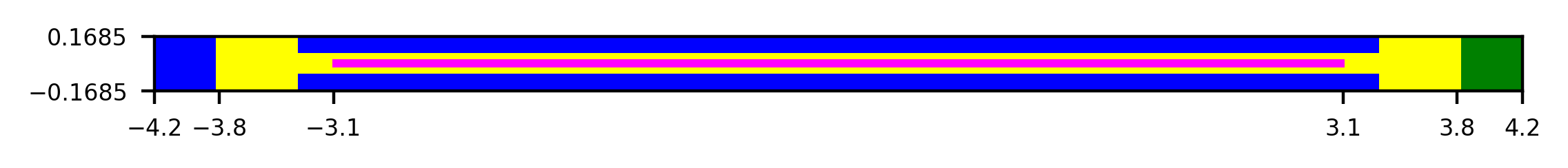}
    \caption{Top view of the system: the colors pink, yellow, blue, and green stand for enriched uranium, aluminum, water, and cadmium, respectively. The numbers are expressed in centimeters, the z-axis is limitless,  and all sides have reflective boundary conditions.}
    \label{fig:slab}
\end{figure}

The system consists of a repeated unit cell describing an MTR fuel assembly with fuel plates of $93\%$ enriched uranium with aluminum cladding, and water cooling in between. perpendicular aluminum side plates hold the plates together. One side plate borders with water, while the other borders with a cadmium absorber plate, as depicted in \autoref{fig:slab}. There is a reflective boundary condition on the x-y sides and the z-axis is infinite.

The variable parameters in this model are the uranium and water densities. We recognize that changing the water density is analogous to adjusting the spacing between fuel plates, but makes it easier to model, as the geometry does not change.

\subsection{Objective Function} 
One of the applications of a reactor is to test materials under high neutron flux. This inspired us to design this MTR benchmark to maximize the fast neutron flux (above 0.6 eV) while still staying in the critical region. Therefore we define the objective function to be:
\begin{equation}
\frac{|k-1|+1}{\phi+1}  
\end{equation}

where $\phi$ is the fast neutrons flux (per neutron source) in the water gap opposite to the cadmium absorber (left of the plate in \autoref{fig:slab}) and $k$ is the multiplication coefficient. It is straightforward to see that the parameters that lead to the smallest fitness value are consistent with the requirements above. The objective function coefficients were determined with the aid of our simulations. Namely, to change the significance of deviation from criticality, we may change the added constant in the numerator. A positive constant was added to the flux in the denominator to avoid singularities. However, these two numerical choices were chosen by trial and error, to reflect the constraints and the obtained fitness values, and changing them provides degrees of freedom to sculpt the fitness map (see the maps in the Results section). Thus, this benchmark lays a foundation to further study the choice of the objective function, as other values may lead to faster convergence or better suitability to the stochastic nature of the MC results.

\subsection{Results}
We let the OpenNeoMC run the MTR benchmark with two optimization algorithms.  The results show a considerable distinction between them. \autoref{result_table} presents the best parameters each algorithm found and their resulting performance. It shows that the neuroevolutionary algorithm (PPO-ES) obtained substantially better results than the evolutionary algorithm (JAYA), leading to a factor of about 300 in the resulting flux. 

\begin{figure}[!ht] 
    \subfloat{{\includegraphics[width=6.5cm]{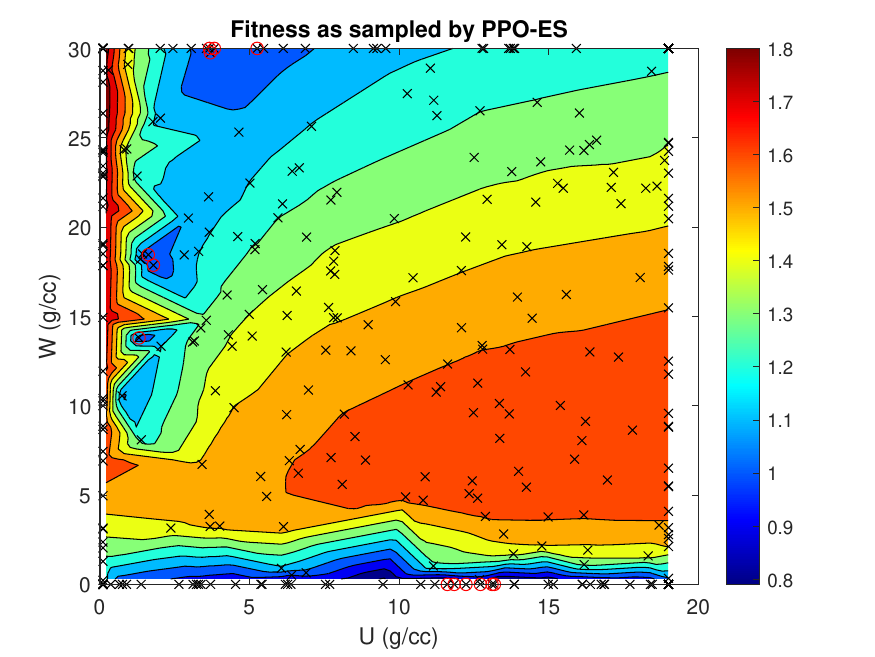} }}%
    \ 
    \subfloat{{\includegraphics[width=6.5cm]{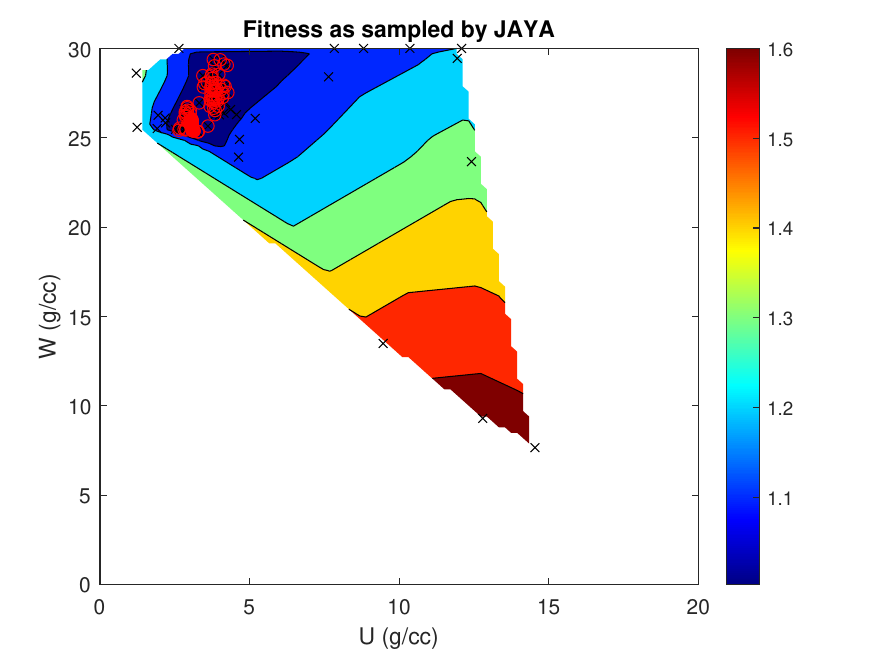} }}%
    \caption{The interpolated fitness map as sampled by the PPO-ES (left) and JAYA (right) algorithms, as a function of the uranium and water density ($U$ and $W$, respectively). The black crosses indicate sampled points, whereas red circles indicate criticality (sample points where $k~1$). JAYA sampling reveals only a subspace of the parameter domain, while PPO-ES maps the full permissible extent.}%
  \label{fig:maps}

    \end{figure}

Each algorithm converged to its results at a comparable number of steps (a few hundred), and the JAYA algorithm we used did not find the better solution even after many more calculation steps. To better understand this behavior it is worth looking at the panels of \autoref{fig:maps} which feature the fitness value landscape as a function of the water and uranium density, as interpolated from the sampled points of each algorithm. As shown, the fitness landscape is not trivial, featuring both a shallow valley with a local minimum and a deep and steep global minimum at the edge of the parameter domain (at zero water density). The local minimum is obtained by the JAYA algorithm, which is totally unaware of the existence of a low fitness area near the domain edge. The global minimum is obtained by PPO-ES, which samples the whole domain. 

\begin{figure}[!ht]
    \centering
    
        \includegraphics[width=0.8\linewidth]{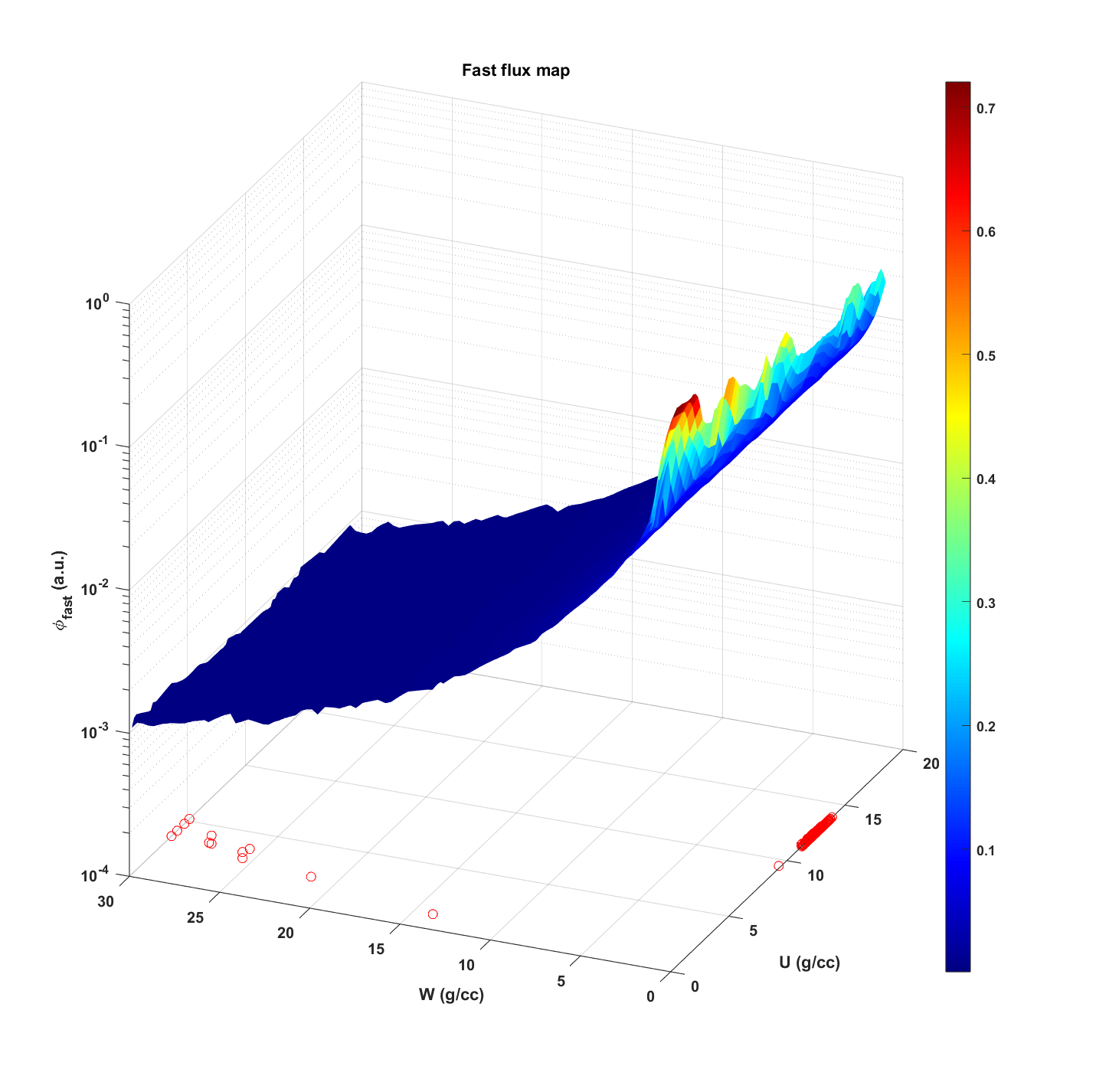}
        \caption{Fast flux in dependence of water and uranium density, the red circles at the base plane indicate areas where criticality is met ($k~1$). The fast flux changes dramatically with water density due to the neutron moderation. The two critical areas are disconnected, preventing genetic algorithms to hop from one to another. Therefore, the JAYA algorithm optimized for high water density, even though the flux there is orders of magnitude lower than at the other critical area.}
   \label{fig:flux}

\end{figure}

The results are detailed in \autoref{result_table}, and the fast flux map is shown in \autoref{fig:flux}. For easier orientation, the regions where $k$ is close to unity are marked by red dots. The physical interpretation of the results is that there are two regimes. The first is the typical thermal reactor regime, where the fission neutrons are moderated to thermal energies and then induce more fissions. The second regime is the fast critical assembly regime, with high uranium density and practically no moderating water, leading to nearly no thermal flux and high fast flux, which induces some fissions without being absorbed by the cadmium. The fast regime is the one the PPO-ES found.

\begin{table}
    \centering
     \caption{Results of the two models, optimized parameters. Each algorithm locked on a different minimum of the fitness in the U-W density domain. The criticality constraint ($k=1$) is met in both cases, but the fast flux obtained by PPO-ES is orders of magnitude higher than that of JAYA.}
   \begin{tabular}{|p{1.5cm}|p{2cm}|p{2cm}|p{2cm}|p{1.5cm}|}\hline 
        Model & Uranium Density (g/cc)& Water Density 
        (g/cc)& Fast Flux (a.u./source)& k\\ \hline 
        JAYA & 2.905 & 24.94 & 0.00141 & 1.000021\\ \hline 
        PPO-ES & 12.23 & 0.001 & 0.4966 & 0.990\\ \hline 
    \end{tabular}
    \label{result_table}
\end{table}

\section{Summary and Conclusions}
In this work, we presented a novel benchmark problem within the OpenNeoMC framework, demonstrating the effectiveness of reinforcement learning in navigating complex optimization landscapes with multiple local minima. The novelty of our benchmark lies in its unique feature of having two different local minima, which poses a challenge for traditional evolutionary algorithms. Through extensive simulations utilizing JAYA and PPO-ES algorithms, we observed that the reinforcement learning algorithm successfully identified the global minima, while the evolutionary algorithm missed it, leading to a far inferior optimization solution.
The introduced MTR benchmark has a simple geometry \cite{openneomc-git} and physical relevance, and yet, the complex fitness landscape makes it difficult for some learning algorithms to converge to the best solution. This benchmark also provides a test case where different objective function parameters and algorithmic hyperparameters may be analyzed to adapt them to the challenges reactor optimization poses when using Monte Carlo simulations.

Our findings highlight the importance of considering diverse optimization scenarios in nuclear reactor engineering, where traditional evolutionary algorithms may fall short in capturing the global optima. By incorporating reinforcement learning techniques, we can effectively address such challenges and obtain more robust and optimal solutions for reactor optimization tasks.

Furthermore, our study contributes to advancing methodologies for addressing intricate optimization challenges in nuclear engineering, paving the way for more efficient and effective reactor design and operation.

\section*{Acknowledgements}
This research was supported by the Pazy Foundation and the Lynn and William Frankel Center for Computer Science. Computational support was provided by the NegevHPC project~\cite{negevhpc}.

\bibliography{bibTex}

\end{document}